\documentclass[letterpaper, 10 pt, conference]{IEEEtran}  
\usepackage{tikz}
\usepackage[mode=buildnew]{standalone}
\usepackage{graphicx}
\usepackage{times} 
\usepackage{amsmath} 
\usepackage{amssymb} 
\usepackage[numbers]{natbib}

\usetikzlibrary{fit,positioning}
\usepackage[english]{babel}

\usepackage{dsfont}
\usepackage{algorithm}
\usepackage{algcompatible}

\usepackage{pgfplots}
\usepackage{pgfplotstable}
\usepgfplotslibrary{groupplots}
\usepackage{tikzscale}
\usetikzlibrary{calc,shapes,arrows,positioning,fit}
\usetikzlibrary{intersections}
\usetikzlibrary{backgrounds}

\usepackage{subcaption}
\usepackage{hyperref}
\hypersetup{bookmarksopen,bookmarksnumbered,
pdfpagemode=UseOutlines,
colorlinks=true,
linkcolor=blue,
anchorcolor=blue,linktocpage,
citecolor=blue,
filecolor=blue,
menucolor=blue,
urlcolor=blue,
}

\definecolor{Observed}{rgb}{0,0,0.0}
\definecolor{Hybrid}{rgb}{0.25,0.4,1.0}
\definecolor{Switching}{rgb}{1.0,0.4,0.0}
\definecolor{EKF}{rgb}{0.4,0.8,0.4}
\definecolor{GP}{rgb}{1.0, 0.0, 0.5}
\definecolor{training}{rgb}{0,0,0.0}
\definecolor{idzero}{rgb}{0,0,0}
\definecolor{idone}{rgb}{0,1,0}
\definecolor{idtwo}{rgb}{1.0,0.4,0.0}
\definecolor{idthree}{rgb}{1,0,1}
\definecolor{idfour}{rgb}{0,1,1}
\definecolor{resetfrom}{rgb}{0.6,0.4,0.8}
\definecolor{resetto}{rgb}{1.0, 0.0, 0.5}

\usepackage{url}

\DeclareMathOperator*{\argmax}{arg\,max}

\newcommand{\figref}[1]{Figure~\ref{#1}}

\newcommand\Mark[1]{\textsuperscript{#1}}

\title{Unsupervised Learning for Nonlinear PieceWise Smooth Hybrid Systems\\[1.5ex]
  {\normalfont\large 
    Gilwoo Lee\Mark{*},  Zita Marinho\Mark{**}, Aaron M. Johnson\Mark{**},\\[-1.5ex]
    Geoffrey J. Gordon\Mark{**}, Siddhartha S. Srinivasa\Mark{*}, and Matthew T. Mason\Mark{**}}
    \\[-1.0ex]
}

\author{
    \IEEEauthorblockA{%
        \Mark{*}
        Paul G. Allen School of Computer Science \& Engineering\\
        University of Washington\\
        Seattle, WA 98195%
    }
    \and
    \IEEEauthorblockA{%
        \Mark{**}Robotics Institute\\
        Carnegie Mellon University\\
        Pittsburgh, PA 15213%
    }}

\begin{document}
\maketitle

\begin{abstract}
This paper introduces a novel system identification and tracking method for PieceWise Smooth (PWS) nonlinear stochastic hybrid systems.
We are able to correctly identify and track challenging problems with diverse dynamics and low dimensional transitions. 
We exploit the composite structure system to learn a simpler model on each component/mode. We use Gaussian Process Regression techniques to learn smooth, nonlinear manifolds across mode transitions, \emph{guard-regions}, and make multi-step ahead predictions on each mode dynamics.
We combine a PWS non-linear model with a particle filter to effectively track multi-modal transitions.
We further use synthetic oversampling techniques to address the challenge of detecting mode transition which is sparse compared to mode dynamics.
This work provides an effective form of model learning in complex hybrid system, which can be useful for future integration in a reinforcement learning setting.
We compare multi-step prediction and tracking performance against traditional dynamical system tracking methods, such as EKF and Switching Gaussian Processes, and show that this framework performs significantly better, being able to correctly track complex dynamics with sparse transitions.
\end{abstract}

\section{Introduction}
\label{sec:intro}
 Having a good model of the interaction between a robot and the environment is critical in reinforcement learning, as it depends on expected reward based on the probability of predicted trajectories~\cite{kober2012reinforcement}. Poor models lead to false approximations of those probabilities, which in turn lead to poor policies. In receding horizon control we also rely on good model estimates to compute locally optimal policies within short timeframes~\cite{murray2009optimization}. While reinforcement learning and optimal control are active fields of research, most existing work assumes that the underlying dynamics is smooth, and application to Hybrid Systems has been limited\cite{posa2013direct, tassa2010stochastic}. In the context of robotics, deterministic hybrid models are often used for locomotion~\cite{johnson2016hybrid},  object manipulation. Although some non-parametric, data-driven approaches to estimate hybrid systems have been proposed \cite{kroemer2015towards, levine2016end}, they assume overly simplified simplified models such as Gaussian Mixture Model or neglect the sparsity of mode transitions in the data.
 
In the field of hybrid systems, many system identification models has been studied for hybrid systems, but most of them are focused on Switched Systems~\cite{lauer2008switched, lauer2010nonlinear,lauer2014piecewise,bloch2011reduced}, in which mode transition is independent of continuous state. These approaches for Switched Systems often perform poorly on PWS-HS, and identification of PWS-HS has been limited to those with affine dynamics for subsystems or linearly separable subsystems~\cite{santana2015learning}.  Others assume system dynamics can be provided in a closed form solution \cite{nunez2014hybrid}, which is limited in its application when we need to learn complex systems with little prior knowledge about the underlying models.

The main contribution of this paper is to present a unsupervised, nonparametric framework that can learn \emph{nonlinear} PWS-HS systems with little prior knowledge of the underlying subsystems. Doing so is challenging in two ways. First, the domains of the subsystems are unknown, which makes mode assignment and function approximation difficult. Second, mode transitions happen very rarely compared to the non-transitioning counterparts, which makes identification of mode transition condition challenging. In order to address these challenges, we heavily exploit the structure of PWS-HS, i.e. piecewise smoothness and the existence of guard regions, which allow us to use simple non-parametric clustering such as Spectral Clustering\cite{von2007tutorial} and function regression methods, such as Gaussian Processes\cite{rasmussen2002infinite}. To address the challenge in the sparsity of transition points, we use oversampling techniques in the transition region. By iterative mode assignment and subsystem estimation, our system converges to a close approximation of the underlying hybrid system. 

The rest of this paper is structured as following. First we define our problem and explain PWS-HS. Then we introduce how our learning system of PWS-HS, and how it uses a particle filter to predict future observations and track them. We evaluate our system in two experiments, in which we compare it with existing methods such as Single GP, Switching GPs\cite{chen2009switching}, and Extended Kalman Filter~\cite{thrun2005probabilistic}.

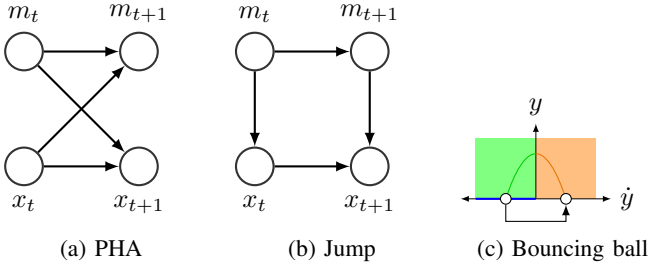
\begin{figure}[t!]
\begin{subfigure}[b]{0.15\textwidth}
\begin{tikzpicture}
\tikzstyle{main}=[circle, minimum size = 5mm, thick, draw =black!80, node distance = 10mm]
\tikzstyle{connect}=[-latex, thick]
\tikzstyle{box}=[rectangle, draw=black!100]
  \node[main, fill = white!100] (m1) [label={above}:$m_t$] { };
  \node[main] (m2) [right=of m1,label={above}:$m_{t+1}$] { };
  \node[main] (x1) [below=of m1,label={below}:$x_{t}$] { };
  \node[main] (x2) [below=of m2,label={below}:$x_{t+1}$] { };
  \path 
        (m1) edge [connect] (m2)
        (m1) edge [connect] (x2)
        (x1) edge [connect] (x2)
        (x1) edge [connect] (m2);
\end{tikzpicture}
\caption{PHA}\label{fig:PHA}
\end{subfigure}
~ \begin{subfigure}[b]{0.15\textwidth}
  \begin{tikzpicture}
\tikzstyle{main}=[circle, minimum size = 5mm, thick, draw =black!80, node distance = 10mm]
\tikzstyle{connect}=[-latex, thick]
\tikzstyle{box}=[rectangle, draw=black!100]
  \node[main, fill = white!100] (m1) [label={above}:$m_t$] { };
  \node[main] (m2) [right=of m1,label={above}:$m_{t+1}$] { };
  \node[main] (x1) [below=of m1,label={below}:$x_{t}$] { };
  \node[main] (x2) [below=of m2,label={below}:$x_{t+1}$] { };
  \path 
        (m1) edge [connect] (m2)
        (m1) edge [connect] (x1)
        (m2) edge [connect] (x2)
        (x1) edge [connect] (x2);
\end{tikzpicture}
\caption{Jump}\label{fig:Switched}
\end{subfigure}
~ \begin{subfigure}[b]{0.15\textwidth}
\includegraphics{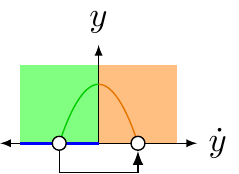}
\caption{Bouncing ball}\label{fig:Ball}
\end{subfigure}
\caption {Probabilistic Hybrid Automata PHA (left). Switched System (middle). model structure: $m_t$ is a discrete mode, $x_t$ is a continuous state undergoing dynamics defined by $m_t$. Probability of mode transition at time $t$ is defined both by the current mode $m_t$ and continuous state $x_t$.}
\label{fig:models}
\end{figure}

\section{Problem Statement}\label{sec:problem}

The goal of our framework is to learn a nonlinear PWS-HS from data when little prior model knowledge is available.  

\subsection{PieceWise Smooth Hybrid Systems}

A state in a PWS-HS is defined as a tuple $(m,x)$. 
Here, $m\in M$ is a \emph{mode} chosen from a set of discrete modes $M$.
Each mode is associated with a continuous, connected subspace $F_{m} \subseteq\mathbb{R}^d$,
and $x \in F_{m}$ is an element of that subspace.
We assume the subspaces do not overlap, so $F_m \cap F_n = \emptyset$ if $m \neq n$.

Each mode $m$ is also associated with a discrete-time continuous function $f_m \in \mathcal{F}: F_m \rightarrow \mathbb{R}^d $, which determines the evolution of $x$ within the mode $m$.

We also define a \emph{guard region}, denoted as $G_{m,m'} \subseteq F_{m}$ in which transition occurs deterministically from mode $m$ to a different mode $m'$. When $x_t$ in mode $m$ reaches $G_{m,m'}$, it teleports to $x_{t+1}$ through a \emph{reset function} $r_{m, m'} \in \mathcal{F}$. 

In short, the continuous state propagates in the following form:
\begin{align}\label{eq:system}
   x_{t+1}&=\begin{cases}
    r_{m_t, m_{t+1}}(x_t) + w_t   &\text{if } x_t\in G_{m_t, m_t'}\\
   f_{m_t}(x_t) + w_t    &\text{if }x_t \in F_{m_t} 
   \end{cases}
\end{align}
where $w_t \sim \mathcal{N}(0, \sigma_w)$ is additive process noise. Although our model does not explicitly model control input $u$, it is straightforward to incorporate. 

Consider, for example, a perfectly elastic bouncing ball (\figref{fig:Ball}). 
The ball undergoes a nonsmooth transition of state when it hits the ground and its velocity changes sign instantaneously.
We can describe this system as having two modes, with transition between the modes upon collision.\footnote{We can interpret this model as having a single mode with self-reset at the bouncing point. For simplicity, we disallow self-resets in our model and view this as two modes with one mode transition.}\\
The bouncing ball clearly illustrates our challenge. Although the dynamics of each mode can be easily learned, it is challenging to learn the guard region, since these are typically low dimensional or sparse. In this case the switching point is instantaneous and the probability of a discrete time step hitting exactly the bouncing moment is approximately zero. Due to this challenging property of mode transitions, many existing SHS identification tools often fail to learn the mode transitions or handle it incorrectly.  This in turn makes the model fail to make any reasonable prediction near mode transitions. In this work we propose to exploit the structure of PWS systems, and design a framework that addresses this challenge. 

We now formally state our problem:
\begin{quote}
We observe the continuous state with some additive noise $\epsilon_t \sim \mathcal{N}(0,\sigma_\epsilon)$ in observation.
We then collect a set of $n$ trajectories, with each trajectory defined as $X_i=\{x_1^i, ..., x_{T_i}^i\}$.
Our goal is to learn a PWS-HS, i.e., the number of modes, the smooth system dynamics within a mode, and the guard regions and reset function for each transition.
\end{quote}

\section{Learning PWS Hybrid Systems}\label{sec:sysid}

Assuming that the number of modes is given, learning a PWS Hybrid System can be decomposed into:
\begin{enumerate}
\itemsep0em 
\item \textbf{Mode identification}: Determine subsets of state space governed by each mode.
\item \textbf{Learn a model per mode}: Estimate the dynamics of each mode.
\item \textbf{Guard estimation}: Identify mode transition regions.
\item \textbf{Learn a reset mapping}: Estimate reset dynamics for mode transitions.
\item \textbf{Iteratively mode assignment}: Compute MAP estimates of mode for each observed dynamics.
\item \textbf{Classification}: Learn classifiers for each mode prediction and transition.
\end{enumerate}
\begin{figure}[t!]
\begin{center}
\includegraphics{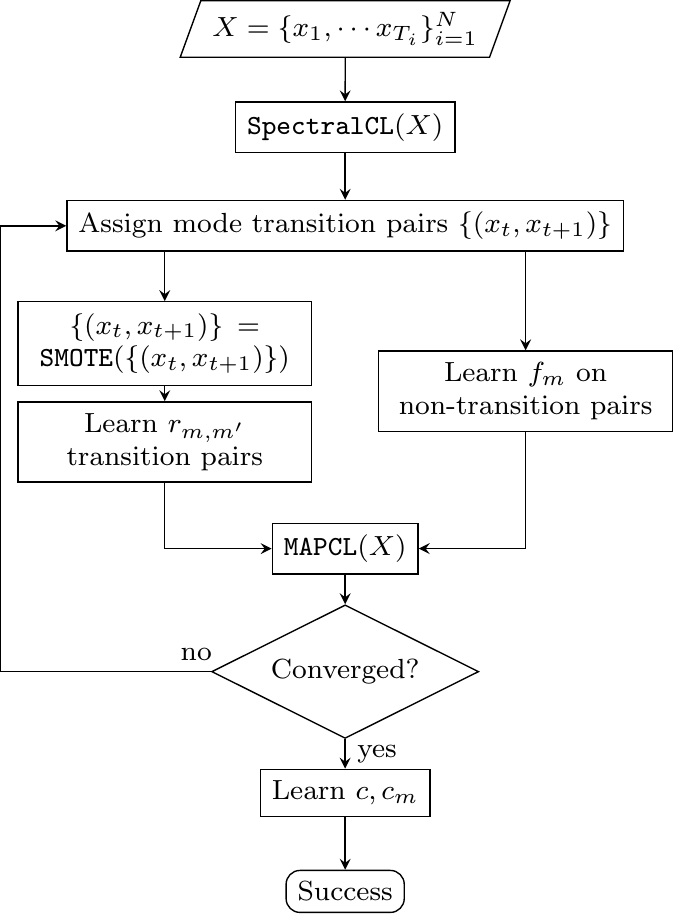}
\caption{Flowchart for system identification and learning \S\ref{sec:sysid}. }
\end{center}
\label{fig:SHS}
\end{figure}

Our approach is focused on constructing a model that closely matches the definition of a PWS-HS, including explicit mode of guard regions and reset dynamics, which many existing approaches fail to address. By doing so, we are able to leverage the structure of the model and learn more effectively with less prior information and data. We learn each component and each transition separately, and hope to achieve a more accurate PWS hybrid model. We design an iterative process that predicts maximum a posteriori mode estimate for each point in the dataset, in a fully unsupervised approach. \\
We start with $n$ trajectories, with each trajectory defined as $X_i=\{x_1^i, ..., x_{T_i}^i\}$. We first assign modes to each trajectory of points using a spectral clustering algorithm (\S\ref{sec:cluster}). Then, for each sequence, we identify mode transition pairs and oversample them (\S\ref{sec:guardid}). We then learn both mode dynamics (\S\ref{sec:hyblearn}) and the reset mapping (\S\ref{sec:gpreset}) through Gaussian Process (GP), and reassign the modes based on the predictions of each GP (\S\ref{sec:iterate}). Once this process converges, we train the mode and guard classifiers using Logistic Regression (\S\ref{sec:classification}). In the following section, we describe the details of this procedure. 

\begin{figure}[t!]
\centering
\includegraphics{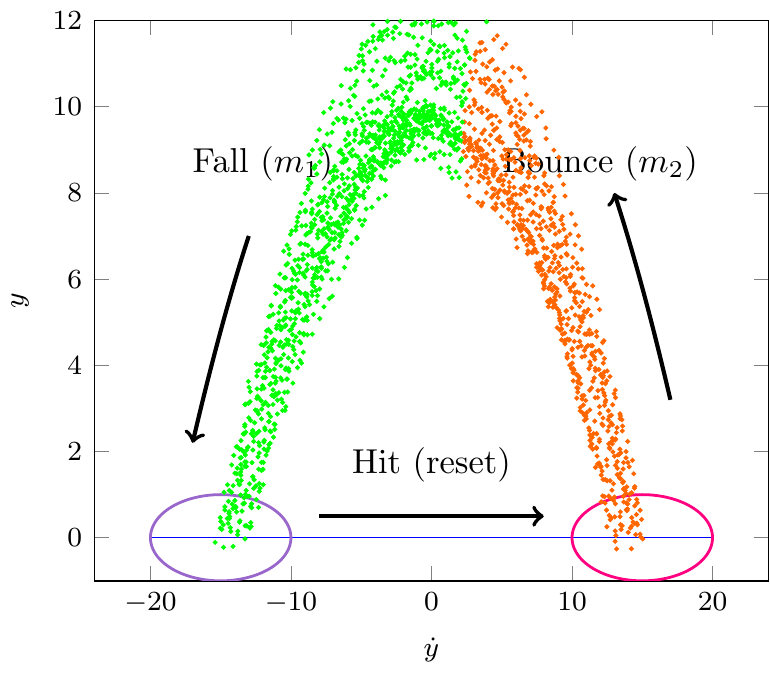}
\caption{Initial mode assignment (\S\ref{sec:cluster}) for bouncing ball trajectories (\S\ref{sec:ball}). The ball falls~($m_1$), reaches the \textcolor{resetfrom}{guard region}, jumps to \textcolor{resetto}{post-guard region}, and bounces up~($m_2$).}
\label{fig:balltraining}
\end{figure}

\subsection{Initial Mode Assignment} \label{sec:cluster}
Spectral Clustering (SC) is a powerful clustering technique that can be used for clustering nonlinear, smooth manifolds~\cite{Ng01onspectral}. We assume each mode dynamics is smooth and non-intersecting. In this case SC serves as an effective initial mode assignment tool in our model. Refer to \cite{von2007tutorial}, \cite{Jordan03} for more detail. 
Given a set of points $X = \cup_{i=1}^n X_i$, we build an affinity matrix using a metric of our choice, in our case given by a Radial Basis Function kernel.\footnote{$k(x_i, x_j) = \exp(-\beta_0 ||x_i - x_j ||^2)$} From this affinity matrix, we apply K-means algorithm to get the mode assignment.\footnote{We assume the number of modes is known a priori, but it can be estimated using an eigengap heuristic~\cite{soltanolkotabi2014,Mohar1997}.} The result is a set of ordered pairs of a continuous state and its respective mode $\tau_i=\{(m_1^i,x_1^i),\cdots, (m_{T_i}^i,x_{T_i}^i)\}$ for each sequence $X_i$. 

Figure~\ref{fig:balltraining} shows the initial clustering result given by SC with the number of modes set to~2. The algorithm clusters most of the falling trajectory as one mode and the bouncing trajectory as another. Note that at the top where the ball changes its velocity from positive to negative, the mode transition is smooth, and thus mode assignment in this region is slightly different every time we cluster. This different initialization does not affect the rest of our framework, as we only require the dynamics within each mode to be smooth.

\subsection{Guard Identification}\label{sec:guardid}
After mode assignments are complete, we identify mode transitions. We create pairs $(x_t, x_{t+1})$ along all trajectories $X_i$ and create an augmented matrix 
\[
X_i^{-+} = \begin{bmatrix} X_i^- & X_i^+ \end{bmatrix} = \begin{bmatrix}
	 x_1^i & x_2^i \\
	 \vdots&\vdots\\
	 x_{T_i-1}^i &x_{T_i}^i
\end{bmatrix} 
\]
Note that from the previous step of mode assignment, each of these points has assigned modes $(m_t^i, m_{t+1}^i)$. Hence, for each mode $m$, we collect all pairs with the same mode $(m_t^i = m, m_{t+1}^i = m)$ in $\hat{F}^{-+}_m$, and collect all pairs with different modes $(m_t^i= m, m_{t+1}^i =m')$  in $\hat{G}^{-+}_{m,m'}$. For example, in figure~\ref{fig:balltraining}, all green points are in $m_1$ and all orange points are in $m_2$, and all pairs of points from $m_1$ to $m_2$ are in $\hat{G}^{-+}_{m,m'}$. At the end of this step, we have $\hat{F}^{-+}_m$, $\hat{G}^{-+}_{m,m'}$ for all modes and mode transitions.

\subsection{Learning Mode Dynamics through Gaussian Process Regression}\label{sec:hyblearn}

From the paired points in $\hat{F}^{-+}_m$, we can learn the dynamics $f_m$ of subsystem $m$.  Recall that $\hat{F}^{-+}_m$ is a set of tuples of $(x_t, x_{t+1})$, both in mode $m$. Let $\hat{F}^-_m$ be the first column of $\hat{F}^{-+}_m$, and $\hat{F}^+_m$ be second column. Now we use Gaussian Processes (GP)~\cite{Rasmussen:2005:GPM:1162254} to learn the mapping $\hat{F}^{-}_m \rightarrow \hat{F}^{+}_m $. A GP models the conditional probability of $F^{+}_m$ given $F^{-}_m$ as a normal distribution. Once the GP is trained, given a new test point $\tilde{x}_t$ in mode $m$, it produces the probability distribution of $\tilde{x}_{t+1}$ as a normal distribution $\mathcal{N}(\mu_m(\tilde{x}_t),\Sigma_m(\tilde{x}_t))$, with parameters: 
\begin{align}\label{eq:GPsingle}
\begin{split}
\mu_m(\tilde{x}_t) &= K(\tilde{x}_t, \hat{F}_m^-)^\top K_{FF}^{-1}\hat{F}_m^+ \\
\Sigma_m(\tilde{x}_t) &= K_{\tilde{x}\tilde{x}}- K(\tilde{x}_t, \hat{F}_m^-)^\top K_{FF}^{-1}K(\tilde{x}_t, \hat{F}_m^-)
\end{split}
\end{align}
where $K(a,b)$ is the Gram matrix defined by the kernel function of our choice\footnote{We use RBF kernel with  additional Kronecker delta term for signal variance: $k(x_i, x_j) = \exp(-\beta_0 ||x_i - x_j ||^2) +\beta_1\delta_{ij} $},  $K_{\tilde{x}\tilde{x}} = K(\tilde{x}_t,\tilde{x}_t)$, and $K_{FF}=K(\hat{F}^-_m,\hat{F}^-_m)$. Throughout the rest of the paper we will denote  discrete mode prediction and continuous dynamics during test time as $\tilde{m}_t, \tilde{x}_t$, while $x_t,m_t$ is reserved for training samples. We choose kernel parameters $\beta=(\beta_0,\beta_1)$ to maximize the likelihood of $P(\hat{F}^+\mid \hat{X}^-; \beta)$. 
See \cite{lawrence2004gaussian} for a full derivation.  

Note that GP is in fact capable of learning non-smooth, discontinuous functions using complex, non-stationary kernels. However, this requires stronger prior knowledge of the underlying model or large amounts of data. On the contrary, by exploiting the structure of PWS-HS, we can use GP with simple kernels such as RBF to learn each mode and combine them to learn a complicated system as a whole. This approach is similar to the Mixture of Gaussian Processes introduced in ~\cite{rasmussen2002infinite}.

\subsection{Learning Reset Dynamics with Oversampling} \label{sec:gpreset}
Ideally, we would like to learn reset dynamics across mode transitions from $G_{m,m'}^{-+}$ in a similar manner as we used $F_{m}^{-+}$ for mode dynamics~(\S\ref{sec:hyblearn}). However, this is challenging because these mode transitions are sparse. For example, in the case of the bouncing ball (figure~\ref{fig:balltraining}) only one mode transition occurs per cycle of ball bounce. To address this challenge, we use a common Machine Learning technique called \emph{oversampling}. 

Extending the Synthetic Minority Over-sampling TEchnique (SMOTE)~\cite{chawla2002smote} algorithm originally proposed for classification, we generate synthetic samples of tuples in $G_{m,m'}^{-+}$. First, we randomly choose two tuples $(x_{t_1}, x_{t_1+1}), (x_{t_2}, x_{t_2+1})$ from $\hat{G}_{m,m'}^{-+}$. Then, we generate a new synthetic point by convex combination of the two tuples:
\begin{equation}\label{eq:smote}
    \begin{bmatrix}x_t\\ x_{t+1} \end{bmatrix} = r \begin{bmatrix} x_{t_1}\\ x_{t_1+1}\end{bmatrix} + (1-r) \begin{bmatrix} x_{t_2} \\ x_{t_2+1} \end{bmatrix}
\end{equation}
where $r$ is randomly chosen from the unit interval $[0,1]$.

 Once enough synthetic tuples are generated, we can learn $r_{m,m'}$ using the same process as \S\ref{sec:hyblearn}, where instead we regress the second column of $G_{m,m'}^{-+}$ on the first column.

\subsection{Iterative Mode Assignment and Model Learning}\label{sec:iterate}
When mode boundaries overlap due to smooth transitions or noise, initial mode assignments may be incorrect for some of the points, especially because we do not take into account time-continuity of points along a trajectory at the initial clustering stage. Similar to~\cite{santana2015learning}, we can fix these mis-assignments by re-assigning modes based on the trained GPs' predictions. After training all GPs, we revisit all points in $X$ and re-assign modes as:
\[
m_t^{i, new} = \argmax_m p(x_{t+1}^i | x_t^i; \mu_{g_m}(x_t), \Sigma_{g_m}(x_t)) 
\]
where $g_m \in  \{f_m\}  \bigoplus \cup_{m'} r_{m,m'} $ is a GP from the set of all dynamics related to mode $m$, i.e., the set of all reset dynamics from $m$ and the mode dynamics $f_m$. We refer to this algorithm as \texttt{MAPCL} in figure~\ref{fig:SHS}. 

We iterate the process of guard identification, oversampling, GP training, and mode assignment until mode assignment does not change.

\subsection{Mode and Guard Classification}\label{sec:classification}
Once the learning process has converged, we finally train classifiers for modes and guards. These can be learned by one classifier, but to explicitly encode the dependency of the current mode and state in the mode prediction for the next time step, we train two level of classifiers. The first level of classification being for mode prediction and then the second level being for guards within each mode. All classifiers return probabilities across possible choices. 

The first level mode classifier is defined as $c:\mathbb{R}^{d}\rightarrow \Delta^{\vert M\vert-1}$, and is trained using all points in $X$ and their assigned modes. The second level classifier is defined per mode: 
\[ 
c_{m}: x \in F_m \rightarrow \Delta^{\vert M\vert-1},
\] 
and we use all points in the guard regions of $m$, i.e. $\cup G_{m,m'}^- \forall m'$, as well as the synthetic points generated in \S\ref{sec:gpreset}. We use balanced Logistic Regression~\cite{fan2008liblinear} for both levels, but the choice of classifier is independent of our framework as long as it provides probabilistic mode estimation and is capable of learning unbalanced data sets.

\section{Multi-step Prediction and Tracking}\label{sec:tracking}
\begin{figure}[t!]
\includegraphics{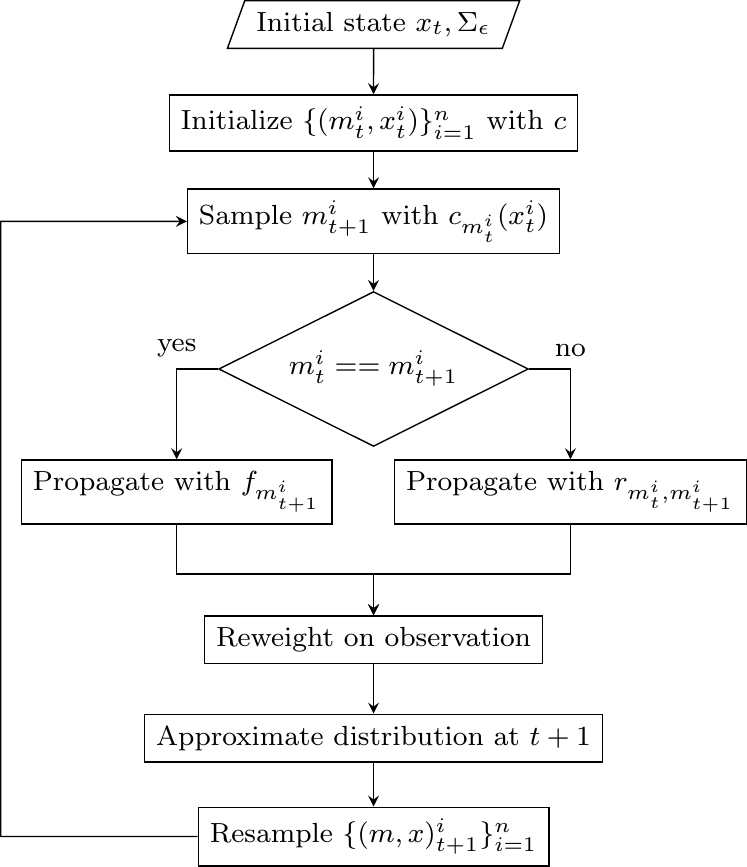}
\caption{Flowchart for tracking with sequential importance sampling (\S\ref{sec:tracking})}
\label{fig:predictionchart}
\end{figure}

Our learned PWS-HS can be used for multi-step prediction and tracking. We combine our learned hybrid model with a Sequential Importance Sampling Particle Filter~\cite{thrun2005probabilistic}. While closed form updates of mean and covariance for multi-step predictions of a single GP are available~\cite{girard2003gaussian}, these are often insuficient in modeling complex systems. In the case of multi-modal PWS-Hybrid Systems closed form updates are often inaccurate, hence we opt for a Monte Carlo approach. 
Figure~\ref{fig:predictionchart} shows a flowchart of our tracking algorithm. For each particle, we first sample the next mode, according to the mode to the learn classifiers, and then propagate the continuous state using the corresponding dynamics function (GP). If an observation is made, it is used to re-weight the particles, and approximate better the target distribution.  Within each mode, we expect the propagation to be unimodal, so we approximate each mode's distribution to be Gaussian, producing a multi-modal Gaussian Mixture Model prediction at the end of each iteration. At the end of the process we resample particles according to the updated mode distribution.

\section{Experiments}
\label{sec:experiments}
We evaluate our hybrid learning system (Hybrid)\footnote{Our system is implemented in Python 2.7.12. For balanced Logistic Regression and Spectral Clustering we used Scikit-learn 0.17.14 \cite{scikit-learn}.} by comparing its prediction and tracking performance with three baseline methods, Single Gaussian Process (GP), a Switching Gaussian Process (Switching) \cite{chen2009switching}, and an Extended Kalman Filter (EKF) \cite{thrun2005probabilistic}. GP assumes that state dynamics is unimodal. Switching assumes that state dynamics is drawn from multiple modes, but assumes that mode evolution is independent of the continuous dynamics. EKF requires knowledge of the model dynamics equation in closed form, and makes mode transitions deterministically based on its state estimation. 

For prediction, we first qualitatively compare $n$-step ahead prediction of our model with all the other baselines. We show that our hybrid learning system is the only method that makes correct predictions near mode transition. We then quantitatively compare the log likelihood of $n$-step ahead predictions and the posterior distribution of predictions given observation for tracking. We show that our model performs as well as the remaining methods, when the state is far from transitioning, and has the best performance near transition. Both GP and Switching use a sequential importance sampling particle filter similarly to our Hybrid model. We use two synthetic experiments as illustrative examples, but we plan to extend it to real robot experiments in the near future. 

\begin{figure*}[ht]
\includegraphics{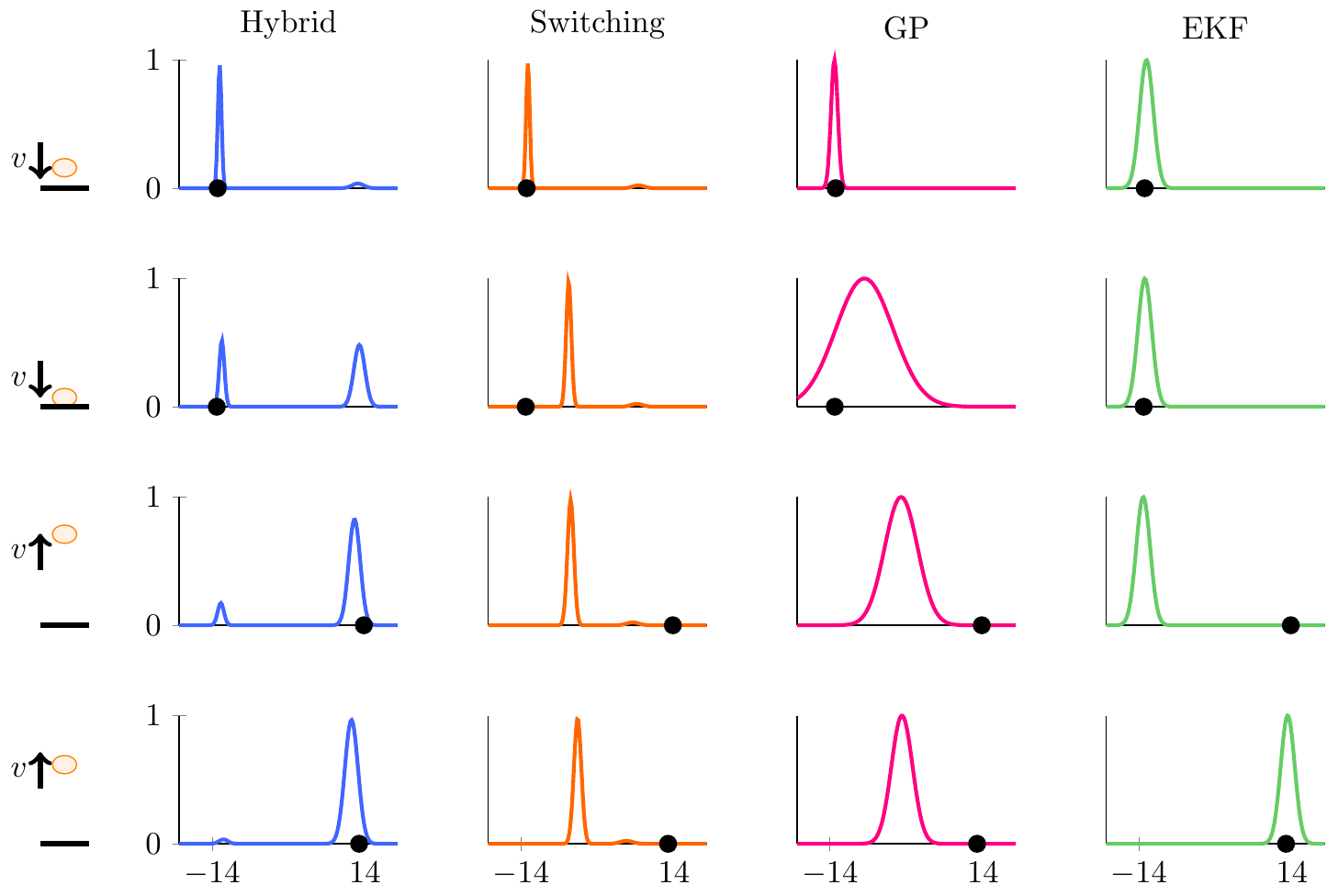}
 \caption{Probability distribution for $v_{t+1}$ prediction, made by Hybrid, Switching, GP and EKF (from left to right). Dot on each x-axis represents the observation. Falling ball (top), immediately before collision (middle top), immediately after collision (middle bottom), ball going up (bottom).}
\label{fig:balltransition}
 \end{figure*}

\subsection{Bouncing ball}\label{sec:ball}
We simulate a one-dimensional bouncing ball experiment. We consider state to be $\begin{bmatrix} y & \hat{y}\end{bmatrix}^\top$, where $y$ is the height of the ball and $\hat{y}$ is the velocity, with positive direction pointing upward. We assume the ground is at $y=0$ and that collisions are perfectly elastic. Except for the collision with ground, the state is updated by the following equation of motion:
\begin{align*}
\begin{bmatrix}y_{t+1} \\ \dot{y}_{t+1}\end{bmatrix} &=
\begin{bmatrix}y_t \\ \dot{y}_t \end{bmatrix} + 
\begin{bmatrix}
      \dot{y}_t t + g t^2/2  \\
      g t
\end{bmatrix} + w_t 
\end{align*}
We add process noise $w_t \sim \mathcal{N}(0, \text{diag}([0.01, 0.01]))$ during the data generation and observation noise of the same magnitude to the training set. We additionally generate 5 trajectories of 100 steps, and use this as test set.
\subsubsection{Prediction and Tracking}\label{sec:predball}
We evaluate tracking performance near the transition region, see \figref{fig:balltransition}. We compare four stages of the bouncing ball experiment out of collision in one mode (ball going down), immediately before collision (guard region), immediately after collision (post-guard), and out of collision in the next mode (ball going upwards). Hybrid correctly predicts that the ball will continue to fall until collision. At this point, the algorithm predicts that either the ball will continue to fall or bounce, with stronger belief for bouncing in the next step. Note that this bi-modal prediction is an accurate representation of how the bouncing ball behaves near collision given noisy observations. This multi-modality is particularly useful in manipulation planning. Once collision is made, the algorithm predicts that the ball will continue to move upward. In all four stages our method correctly predicts the velocity of the ball, giving as well a correct bi-modal distribution at the collision points (guard and post-guard). EKF and GP make unimodal predictions, and Switching makes incorrect multimodal predictions in the post-guard region, not being able to correctly map to the different dynamics in the new mode. 

\begin{figure}
\includegraphics{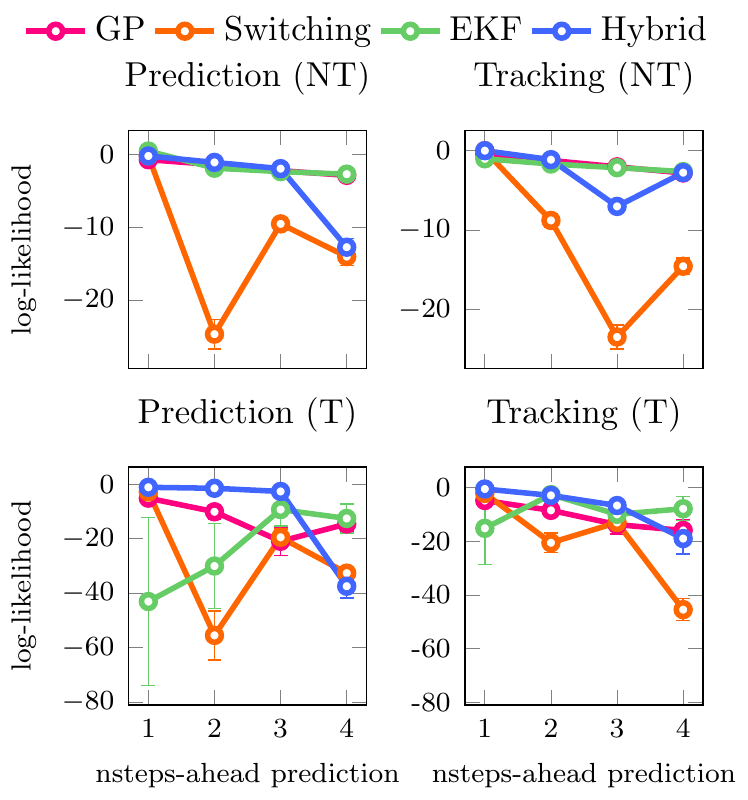}
\caption{Bouncing ball log-likelihood of velocity prediction vs. n-time steps ahead for prediction (left) and tracking (right). Hybrid (blue), GP (red), Switching (orange) and EKF (purple). On each mode subspace--no transition (top), at the transition region (bottom).}
\label{fig:ballllh}
\end{figure}

We also compare the prediction performance of our method, by evaluating the log likelihood of 2-steps ahead prediction and compare with the other baselines, in \figref{fig:ballllh}. 
 We show log-likelihood of observations after updating the probability distribution given the observation. When there is no mode transition, all methods make similar predictions, though Hybrid and Switching perform better than the other two methods, since they learn the system's dynamics of each mode separately. EKF performs best best during this no transition region, which is expected since it is given the correct equations of motion. However, near transition, EKF performs poorly, since the initial prediction is often too far from what is observed. During transition, Switching preforms better in tracking than in pure prediction, see Figure~\ref{fig:ballllh}. This is due to an increased weight for particles in the correct mode after the observation update. Nonetheless, Hybrid performs better than any other methods during mode transition, as none of the others correctly identifies the reset (bouncing).

\subsection{Box pushing}\label{sec:box2d}
\begin{figure}
\centering
\begin{subfigure}[b]{0.47\columnwidth}
\centering
\includegraphics[height=3cm,width=3cm]{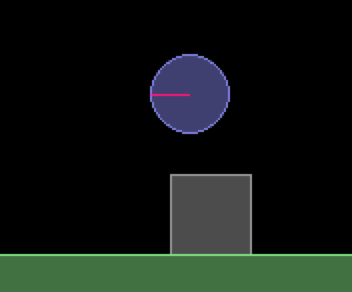}
\caption{Miss the box.}
\end{subfigure}
~
\begin{subfigure}[b]{0.47\columnwidth}
\centering
\includegraphics[height=3cm,width=3cm]{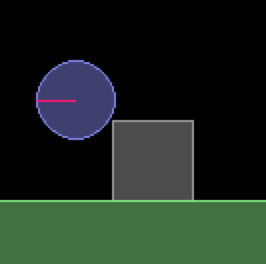}
\caption{Hit the box.}
\end{subfigure}
\caption{The box is dragged by the manipulator $50\%$ of the time, and slows down over until it stops (right). The other half the box remains at rest (left).}
\label{fig:boxpushing}
\end{figure}

In this second experiment, we consider a simplified robot pushing a box, depicted in Figure~\ref{fig:boxpushing}. Here, a circular robot of radius $2cm$, and a box of size $4cm\times 4cm$ sitting 6cm apart from the robot. The robot moves in the positive $x$ direction for $1s$, with velocity $4cm/s$, and stays still for the next $1s$. 30 trajectories were generated for training with additive observation noise $\mathcal{N}(0, 0.1)$, and 6 of them were used for testing. The state space has 5 dimensions, $\begin{bmatrix} x_o, v_{x,o}, x_r, y_r, v_{x, r}\end{bmatrix}$, where $x_o, v_{x,o}$, $x_y, y_r, v_{x,r}$ describe object and robot's pose and velocity, respectively. Both the robot and object move only in the $x$ direction. The robot selects initial height from normal distribution, i.e., $y_r \sim  \mathcal{N}(6cm, 0.5)$, which results in successfully pushing the box half of the time. 

For EKF, we use the following equations of motion:
\begin{align*}
	\begin{bmatrix}
		x_o\\
		x_r\\
	\end{bmatrix}^{(t+1)}
	= 	\begin{bmatrix}
		x_o\\
		x_r\\
	\end{bmatrix}^{(t)} + \begin{bmatrix}
		v_{x,o}\\
		v_{x,r}\\
	\end{bmatrix}^{(t)} {t_{step}}
\end{align*}
\begin{align*}
\begin{cases}
		{v_o}^{(t+1)} = {v_{x,r}} \qquad \text{if }  x_o - x_r \geq  4, y_r \geq 6\\
		{v_o}^{(t+1)} = 0 \qquad  \qquad \text{otherwise}
\end{cases}
\end{align*}

where $t_{step}$ is the step size, 0.2s.

\subsubsection{Prediction and tracking}
\begin{figure}[t!]
	\includegraphics{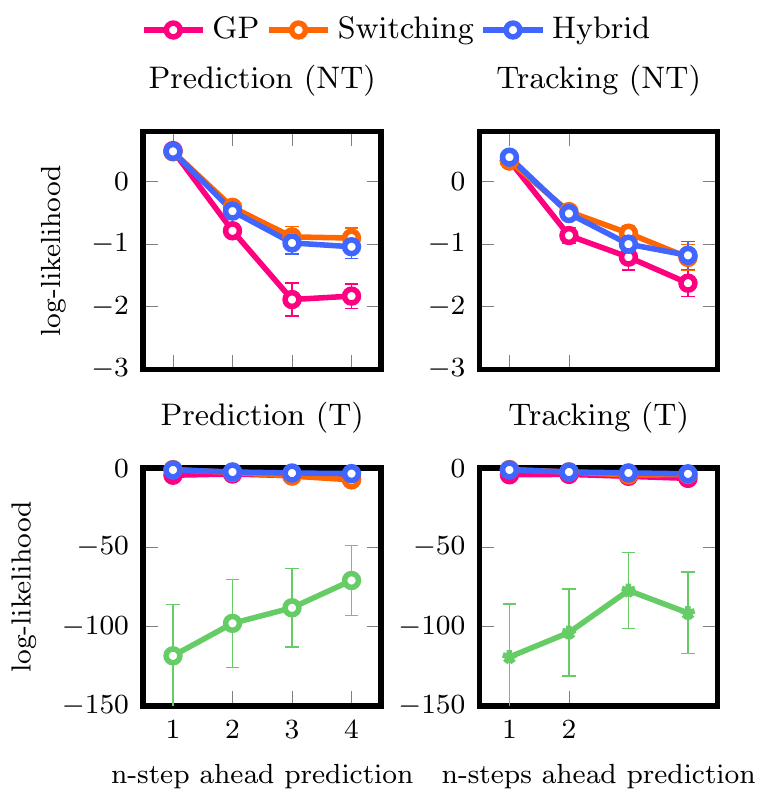}
\caption{Pushing box log-likelihood of velocity prediction vs. n-time steps ahead for prediction (left) and tracking (right). Hybrid (blue), GP (red), Switching (orange) and EKF (purple). On each mode subspace--no transition (top), at the transition region (bottom).}
\label{fig:boxmultisteppredictions}
\end{figure}

Figure~\ref{fig:boxmultisteppredictions} compares the log likelihood of 1 and 2-step ahead predictions made by the four algorithms. As before, we distinguish predictions near transition from the rest. When there is no mode transition Hybrid, GP, and Switching are perform almost equally well. However near transition, Hybrid outperforms the other methods. We do not report EKF's performance since it is much worse than other methods because it's equation of motion is deterministic on what it believes is the true mode, and so it is more prone to errors. 
Moreover, as in the previous experiments, we are able to observe successful multi-modal predictions near transition for this task, similarly to Figure~\ref{fig:balltransition}. These synthetic experiments show the effectiveness of our method, we are currently working on experiments with more realistic robotic manipulation experiments. Nevertheless, with our hybrid system we are able to make multi-modal predictions near transition, capturing the actual distribution of trajectories near mode transitions, and outperforming current state of the art tracking algorithms.

\section{Discussion}
\label{sec:discussion}

We have presented an unsupervised learning framework for PieceWise Smooth Hybrid Systems, which can be used to model the underlying dynamics of many robot systems in which the robot interacts with environment. This work addresses a complex robotic problem and a challenging learn task. We provide an algorithm that is able to learn hybrid dynamics in an unsupervised manner, we exploit the structure of the system to detect guard regions and learn mode and reset dynamics separately, in an iterative process. We have shown through our experiments that this model captures the multimodality of the dynamics near mode transitions, which many existing techniques fail to capture. We believe that having a proper hybrid systems model is crucial in generating policies for robots interacting with the environment, and we are excited to extend this work in such direction in the near future.

\addtolength{\textheight}{-12cm} \bibliographystyle{IEEEtranN}

\bibliography{references}

\end{document}